\pdfoutput=1

\documentclass[11pt]{article}

\usepackage{EMNLP2022}
\usepackage{graphicx}
\usepackage{subfigure}
\usepackage{booktabs} 
\usepackage{xspace}

\usepackage{algorithm}  
\usepackage{algpseudocode}  

\usepackage{CJK} 

\usepackage{multirow}  

\usepackage{hyperref}

\usepackage{amsmath}
\usepackage{amssymb}
\usepackage{mathtools}
\usepackage{amsthm}

\usepackage[capitalize,noabbrev]{cleveref}

\theoremstyle{plain}

\theoremstyle{definition}

\theoremstyle{remark}

\newcommand{\proposed}{GPS\xspace}

\usepackage{pifont}
\newcommand{\cmark}{\ding{51}}%
\newcommand{\xmark}{\ding{55}}%

\usepackage[textsize=tiny]{todonotes}

\usepackage{makecell}

\usepackage{times}
\usepackage{latexsym}

\usepackage[T1]{fontenc}

\usepackage[utf8]{inputenc}

\usepackage{microtype}

\usepackage{inconsolata}

\usepackage{hyperref}

\newcolumntype{M}[1]{>{\centering\arraybackslash}m{#1}}
\newcolumntype{N}{@{}m{0pt}@{}}

\title{GPS: Genetic Prompt Search for Efficient Few-shot Learning}

\author{
Hanwei Xu$^{*}$, Yujun Chen$^{*}$, Yulun Du$^{*}$, \\
{ \bf Nan Shao, Yanggang Wang, Haiyu Li, Zhilin Yang$^{\dagger}$} \\
    Recurrent AI \\
    \texttt{\{xuhanwei, chenyujun, duyulun, kimi\_yang\}@rcrai.com}
}

\begin{document}
\maketitle
\begin{abstract}
Prompt-based techniques have demostrated great potential for improving the few-shot generalization of pretrained language models. However, their performance heavily relies on the manual design of prompts and thus requires a lot of human efforts. In this paper, we introduce Genetic Prompt Search (\proposed) to improve few-shot learning with prompts, which utilizes a genetic algorithm to automatically search for high-performing prompts.
\proposed is gradient-free and requires no update of model parameters but only a small validation set. 
Experiments on diverse datasets proved the effectiveness of GPS, which outperforms manual prompts by a large margin of 2.6 points. 
Our method is also better than other parameter-efficient tuning methods such as prompt tuning.
\end{abstract}

{\let\thefootnote\relax\footnote{{$*$ Equal contribution}}}
{\let\thefootnote\relax\footnote{{$\dagger$ Corresponding author}}}
{\let\thefootnote\relax\footnote{Code is available at \href{https://github.com/hwxu20/GPS}{https://github.com/hwxu20/GPS}}}

\section{Introduction}

Pretrained language models, such as BERT \cite{devlin-etal-2019-bert}, XLNet \cite{yang2019xlnet}, T5 \cite{JMLR:v21:20-074}, and GPT \cite{radf2019lmmultitask}, are often finetuned for downstream natural language processing tasks, which has been shown to improve performance over non-pretrained models. However, this pretraining-finetuning paradigm still relies on a relatively large set of labeled data for each downstream task to obtain competitive performance. 
Although GPT-3 \cite{NEURIPS2020_gpt3} shows promising performance for zero-shot and few-shot learning by prompting on an extremely-large pretrained language models with 175B parameters,
finding out the optimum prompt for each given task could be difficult~\cite{mishra2021reframing,wang2022language}.

To improve the performance of prompting on pretrained language models, recent works focus on supervised pretraining with carefully designed or crowdsourced manual prompts~\cite{eval-harness,wei2021finetuned,sanh2021multitask,Ouyang2022InstructGPT}.
Diverse prompts are collected to enhance the robustness and performance of prompting~\cite{sanh2021multitask}. \citet{Ouyang2022InstructGPT} introduced a dataset of labeler demonstrations and used it to finetune GPT-3.
Despite all these efforts, the challenge to obtain high-performing prompts for few-shot learning still exists. 
As pointed out by previous works \cite{liu2021gpt,gao-etal-2021-making,liu2021pre}, manual prompts are usually suboptimal and suffers a high variance on performance.

\begin{figure*}
    \centering
    \includegraphics[width=0.95\textwidth]{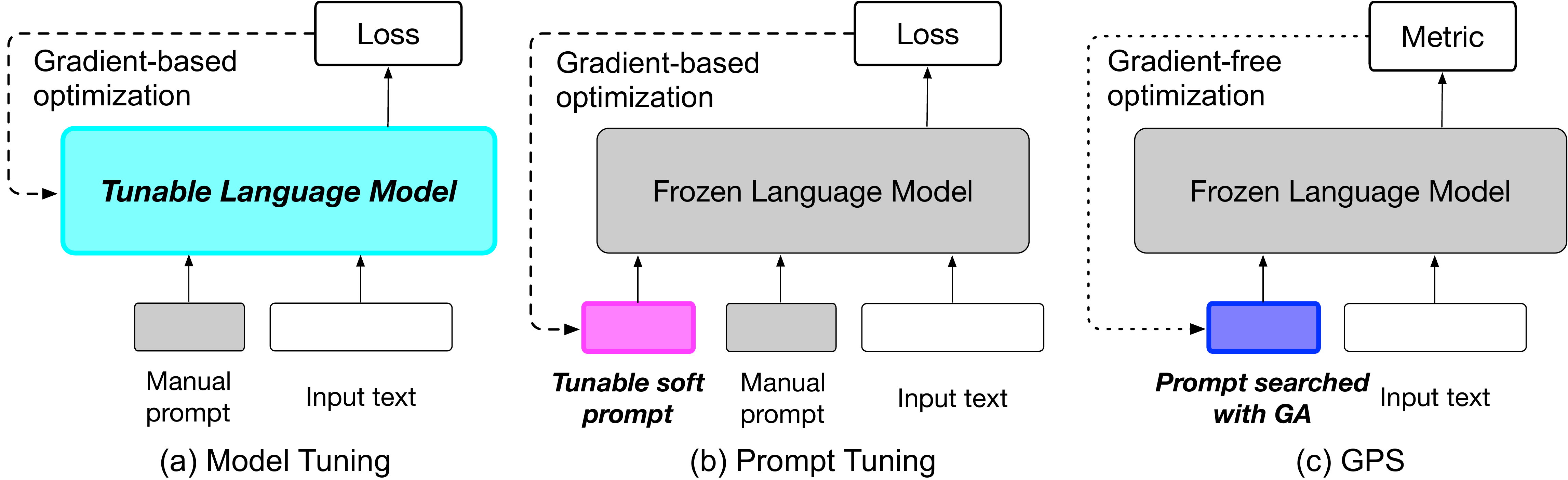}
    \caption{The paradigms of Model Tuning, Prompt Tuning, and \proposed. Model Tuning requires the pretrained model to be tunable, and the tuned model can be only used for a single task. Prompt tuning needs extra tunable soft prompts. Our proposed \proposed is tuning-free.}
    \label{fig:intro_GPS}
\end{figure*}

To address this challenge, we propose a novel Genetic Prompt Search (\proposed) algorithm that gradually mutates the prompts with a generative model and selects candidates according to their performance on a small development set. This evolutionary procedure relies on a tiny set of labeled data, only used for validation but not training.
As illustrated in Figure~\ref{fig:intro_GPS}, GPS does not require updating any parameter, but only searches for the optimal hard prompts for every downstream task.
Similar to prompt tuning, \proposed allows the pretrained model to serve a large number of applications simultaneously. Meanwhile, \proposed is even easier to deploy than prompt tuning, because it does not need to store the tuned continuous soft prompts. 
Empirically, \proposed achieves substantial improvement over the baseline of manual prompts, and it also outperforms other parameter-efficient few-shot tuning methods.

Our contributions can be summarized as follows. 
\begin{itemize}
    \item We propose a tuning-free Genetic Prompt Search method that only requires a small validation set to automatically search for high-performing prompts.
    \item Our experiments demonstrate that manual prompts are usually suboptimal. Using the proposed search method, some simple augmentation technique such as back translation can lead to considerable improvement over manual prompts. Further, our best practice outperforms parameter-efficient few-shot tuning baselines. We conduct an overall comparison of different few-shot learning methods, and our proposed method stands out with the best performance as well as serving efficiency.
    \item We carefully studied the effects of hyper-parameters and prompt generation strategies in the proposed algorithm. Different from previous work~\cite{shin-etal-2020-autoprompt}, we find that our searched prompts are semantically fluent just as human-written templates.
\end{itemize}
\begin{figure*}[!t]
    \centering
    \includegraphics[width=0.99\textwidth]{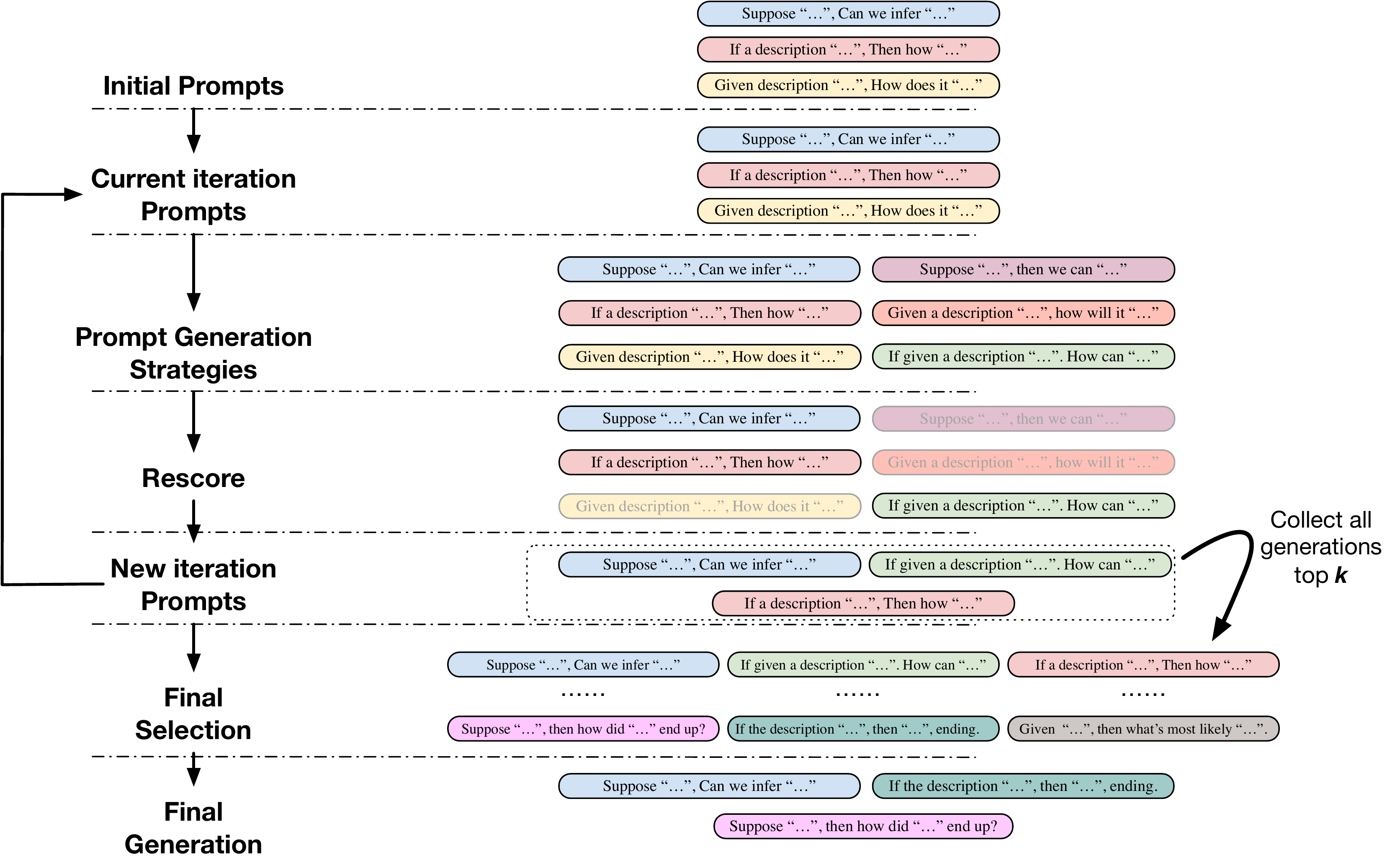}
    \caption{Overall pipeline of our GPS algorithm. The idea of GPS is borrowed from the genetic algorithm. Prompts are initialized from handcrafted prompts. Better prompts are searched for over each iteration. Finally, all generated prompts are reranked and selected as the final prompts. }
    \label{fig:gps_main}
\end{figure*}

\section{Related Work}
Recently, thanks to the prompt-based learning method, pretrained language models (PLMs) have been widely explored under zero-shot and few-shot scenarios for language understanding and generation tasks~\cite{schick-schutze-2021-exploiting,gao-etal-2021-making,le-scao-rush-2021-many}.
Prompt-based learning bridges the gap between pretraining and finetuning objectives by stitching the text input $X$ with a prompt template and augmenting the label output $y$ as a text string, such that the input and the output can be constructed in a sentence completion task form.
Previous few-shot learning methods can be generally categorized into two types, few-shot tuning methods that require updating parameters~\cite{liu2021gpt,han2021ptr} and prompt enchancement methods that have no learnable parameter but optimize the discrete prompts directly~\cite{shin-etal-2020-autoprompt,mishra2021reframing}.

\subsection{Few-Shot Tuning}
Some few-shot tuning methods focus on template design and update all the parameters of pretrained language models. PET\cite{schick-schutze-2021-exploiting} exploited the simple manual template and unified different classification tasks with pattern-verbalizer pairs. LM-BFF~\cite{gao-etal-2021-making} proposed several simple techniques for better few-shot learning including automatic verbalizer search and automatic prompt search. \citet{han2021ptr} applied rules in prompt tuning to deal with the hard many-class text classification tasks.

Another line of work is parameter-efficient few-shot learning, which aims at reducing the number of tunable parameters to improve deployment efficiency. Adapters~\cite{pmlr-v97-houlsby19a} proposed to add an adpater module integrated in the original language model for each downstream task and only this module is tunable. \citet{lester2021power} showed the effectiveness of tuning the prompt embeddings only especially for large-scale models.
P-tuning~\cite{liu2021gpt} applied continuous prompt embedding optimization for GPT and made it comparable to BERT on NLU tasks.
BitFit~\cite{ben-zaken-etal-2022-bitfit} tuned only the bias terms of the original model. A low-rank decomposition approach named LoRA was proposed in \citet{Hu2021LoRA}, which injected trainable matrices in parallel with the original forward pass into each layer.
Black-Box Tuning~\cite{Sun2021blackbox} is a gradient-free optimization method for prompt tuning and thus it is suitable to use language models as a service.
However, all these aforementioned methods require updating parameters, which is computationally expensive and costly in storage capacity for serving every task at hand. Our \proposed, instead, is tuning-free and aims at searching for the optimal prompts.

\subsection{Prompting Enhancement}
GPT-3~\cite{NEURIPS2020_gpt3} shows the effectiveness of In-Context Learning. However, discrete prompting requires human efforts to provide manual prompts, and its sensitivity to labeled examples makes it hard to obtain stable performance.
AutoPrompt~\cite{shin-etal-2020-autoprompt} proposed to search discrete prompts with a gradient-guide method, but the generated prompts are literally uninterpretable. GPTk~\cite{mishra2021reframing} explored several ways to manually reframe task instructions.

GRIPS~\cite{prasad2022grips} is a concurrent work, which also applies iterative prompt search to improve the few-shot performance. Several operations, including add, deletion, swap and paraphrase, are defined to edit the manual prompts. Compared to GRIPS, our method uses different prompt reproduction approaches including back translation as well as cloze and sentence continuation by using generative language models. These methods do not need any human-defined edit rule and the generated prompts are semantically fluent. We conduct experiments comparing the performance of GRIPS and our method in Sec~\ref{sec:exp}.

In this paper, we follow T0~\cite{sanh2021multitask}, which is a very powerful zero-shot baseline of multitask prompted training.
Different from T0 and other methods, we regard the crowd-sourced manual prompts as seed prompts, and focus on parameter-free and gradient-free prompt search to further improve prompting performance under the few-shot learning setting.

\section{Genetic Prompt Search}


In this section, we will introduce the algorithm of Genetic Prompt Search (GPS) and various prompt generation strategies we have studied. Note that the prompt search here refers to the search for a high-performing hard prompt in the discrete word space as shown in Fig.~\ref{fig:intro_GPS}, and the formulation does not include soft prompts.





\subsection{Genetic Prompt Search Algorithm}
\label{subsection:genetic_prompt_search}

It is challenging to automatically find high-performing prompts for a new unseen task.
Inspired by Genetic Algorithms~\cite{mitchell80}, we propose Genetic Prompt Search (GPS) for this purpose.

In \proposed, we will first sample a tiny number of data as a development set $D_{dev}$ for each downstream task. 
Then, we will design two genetic functions, where $f_{GPS}$ is the metric function to decide which prompts will be reserved or eliminated at each iteration, and $g_{GPS}$ represents the genetic function to generate new prompts.
The process of Genetic Prompt Search is described in Fig.~\ref{fig:gps_main}.
According to the algorithm, GPS is firstly initialized with a set of handcrafted prompts, $G^0$.
And the key process of GPS is to reproduce the current generation of prompts and use re-scoring to select prompts iteratively. 
For each iteration, we calculate the scores of prompts in $G^t$ using $f_{GPS}$, and select the top-$K$ prompts as $G^t_*$. 
Then we generate $G^{t+1}$ using $g_{GPS}$ based on $G^t_*$.
After several steps of genetic search, we will collect all the top-$K$ prompts in each generation, and rescore all these prompts to make the final decision on which prompts are optimal.

Now we discuss several strategies to generate the candidates at each iteration.

\subsection{Prompt Generation Strategies}
\label{subsection:prompt_generation}

\begin{algorithm}[t]
  \caption{Genetic Prompt Search}  
  \label{alg:ga-Framwork}  
  \begin{algorithmic}[1]
    \Require  
      $G^0$; $D_{dev}$; $f_{GPS}$; $g_{GPS}$; $T$; $K$;
    \Ensure  
      Final optimized prompts, $G^{T+1}$
    \State obtain handcrafted prompts $G^0$ as initialization
    \For{each $t\in [0,T]$}
        \State store $G^t$
        \State calculate score for each prompt in $G^t$ using $f_{GPS}$,
        \State from $G^t$, select top $K$ prompts as reproductive group $G^t_*$,
        \State generate $G^{t+1}$ based on $G^t_*$ using $g_{GPS}$,
    \EndFor
    \State from stored $\{G^0_*,...,G^T_*\}$, select top $K$ prompts as optimal prompts group $G^{T+1}$ using $g_{GPS}$ . \\
    \Return $G^{T+1}$;  
  \end{algorithmic}  
\end{algorithm}

\textbf{Back Translation}: Back Translation (BT), a common technique for data augmentation in NLP, is applied for prompt reproduction. Here we first translate the manual prompts from English to 11 other languages including Chinese, Japanese, Korean, French, Spanish, Italian, Russian, German, Arabic, Greek, Cantonese, and then translate them back to English.

\textbf{Cloze}: We introduce a prompt generation approach making use of the cloze task form and pretrained language models. Firstly, we follow previous work LM-BFF~\cite{gao-etal-2021-making}, which is a suite of simple techniques for few-shot learning, and exploit its automatic template generation method. Specifically, we use the large pretrained text-to-text transformer (T5)~\cite{JMLR:v21:20-074} to generate templates. For each input example and its verbalizer, we compose the template with placeholders as prefix and suffix, and let T5 to fill in the placeholders. We apply beam search to generate multiple prompt candidates. More details can be found in~\citet{gao-etal-2021-making}. However, this approach does not work well since our setting conducts no parameter update, which is different from the few-shot training setting in the original paper. Therefore, we instead use manual prompts as initial templates, replace some random tokens with placeholders, and then let T5 fill in the blanks to generate new prompts. We select the best prompt according to the average logits across all the validation samples.

\textbf{Sentence Continuation}: Another alternative for prompt augmentation is Sentence Continuation (SC). Inspired by DINO~\cite{schick2020generating},
we use a pretrained language model to generate new prompts. Specifically, we use the template ``Write two sentences that mean the same thing. Sentence 1: \textit{Manual Prompt}, Sentence 2:" to the pretrained model, and let it generate continuations as a new prompt. We conducted experiments with GPT2-XL (1.5B) and T5LM-XXL (11B) as our prompt generation models.

\textbf{Prompt Scoring}:
For Cloze, we follow previous work~\cite{gao-etal-2021-making} to score the prompts with average logits on the validation set $D_{dev}$. For Back Translation and Sentence Continuation, since averaging logits is not applicable, we score each prompt using accuracy on $D_{dev}$.
\section{Experiments}
\label{sec:exp}

In this section, we conduct extensive experiments to study the effectiveness of GPS, and reveal the way to obtain the best prompt for Genetic Prompt Search. We also study several possible impact factors and hyper-parameters in GPS. 

\subsection{Experimental Setups}
To match with the real few-shot scenario, we use a small validation set randomly sampled from each task.
Empirically, for every task, only 32 data samples are needed to build the validation set, and we keep the number of samples for each task the same to make the data balanced. The actual shot number will be 32 divided by the number of classes. For example, we will have 8 shots for each class if there are 4 classes.
Our few-shot setting follows the ``true few-shot'' setting~\cite{perez2021true}. For all the tuning-free methods that do not require a training set, we use the validation set to search for the optimal prompt.
For all the methods that require tuning parameters, we split the validation set into two halves as a training set and a validation set. Therefore all the experiments use the same number of data for fair comparison. We repeat
the experiments of few-shot methods with 3 different data splits and report the average performance across all prompts.

\subsection{Datasets}
We use the 10 test tasks of T0, which are not included in the prompted training tasks, to evaluate the performance of our \proposed and other methods.
There are various kinds of NLP tasks in the test set including natural language inference (ANLI R1, ANLI R2, ANLI R3, CB, RTE), coreference resolution (WSC, Winogrande), sentence completion (COPA, HellaSwag) and word sense disambiguation (WiC). We report the average accuracy of different prompts for all the tasks.

\subsection{Baselines}
We compare \proposed under the few-shot learning setting with state-of-the-art methods. Here we categorize the baselines to three groups: the manual prompt baseline, methods with tunable parameters, and methods without tunable parameters. 

\textbf{Manual prompt baseline}: T0~\cite{sanh2021multitask} is a multitask pretrained encoder-decoder model, which is on the basis of T5 and further pretrained on different types of downstream tasks with diverse manually designed prompts.

\textbf{Methods w. tunable parameters}: 
1) Model Tuning (MT) is the common paradigm to finetune the entire pretrained language model on each task.
2) Prompt Tuning~\cite{webson2021prompt} (PT) is a gradient-guided tuning method, which only trains the extra continuous soft prompts while the pretrained language model is frozen. 
3) Black-Box Tuning~\cite{sun2022black} (BBT) is a gradient-free few-shot tuning method. Rather than searching for discrete text prompts, Black-Box Tuning aims at searching for the best soft prompt embedding in the continuous space.

\textbf{Methods w.o. tunable parameters}: 
1) In-Context Learning~\cite{NEURIPS2020_gpt3} (ICL) is a common method of few-shot learning for large-scale pretrained language models. Demonstrations  composed of labeled samples and manual templates are used to help the model understand the meaning of the test tasks.
2) GRIPS~\cite{prasad2022grips} is a concurrent work which introduced a gradient-free edit-based method for optimal prompt search, but GRIPS mostly focus on simple rule-based editing operations such as add, deletion and swap. 


We conduct experiments on English natural language processing tasks of which the manual prompts are introduced in T0~\cite{sanh2021multitask}. 
Note that all the seed prompts we used in our experiments were taken from T0. To make fair comparison, we used the same suite of prompts for other baseline approaches including Model Tuning, Prompt Tuning, Black-Box Tuning and GRIPS.

For Prompt Tuning, we use the Adafactor Optimizer and set the learning rate as 0.05. 
For Model Tuning, we use the same Optimizer as Prompt Tuning and set the learning rate as 5e-5. The batch size is set as 4 for both prompt tuning and model tuning. 
For Black-Box Tuning, we take 500 as the intrinsic dimension, 20 as the pop size and the cross entropy loss. We report the best results with 1 and 50 soft prompt tokens.
For In-Context Learning, we randomly select 2 examples from the training set for each task.
For GRIPS, we try to keep all hyper-parameters the same as \citet{prasad2022grips}. The only difference is that the initial prompts are from T0.

\begin{table*}[t]
\small
\centering
\begin{tabular}{cm{0.45in}<{\centering}m{0.45in}<{\centering}m{0.45in}<{\centering}m{0.45in}<{\centering}m{0.45in}<{\centering}m{0.45in}<{\centering}m{0.45in}<{\centering}m{0.45in}<{\centering}}
\toprule
\multirow{2}{*}{Dataset} & \multicolumn{2}{c}{Zero-Shot} & \multicolumn{3}{c}{Few-Shot Parameter Tuning} & \multicolumn{3}{c}{Few-Shot Parameter Frozen} \\
\cmidrule(lr){2-3}
\cmidrule(lr){4-6}
\cmidrule(lr){7-9}
& $\text{T0}^\dag$ & T0 $\ddag$ & BBT & PT & MT & ICL & GRIPS & Our GPS \\
\midrule
ANLI R1    & $43.56$ & $43.16$ & $42.97_{0.32}$ & $43.21_{0.15}$ & $\underline{46.73_{3.25}}$ & $37.87_{0.80}$ & $\mathbf{44.41_{0.01}}$ & $44.06_{2.78}$\\
ANLI R2    & $38.68$ & $38.68$ & $38.89_{0.05}$ & $37.36_{1.69}$ & $\underline{39.12_{3.17}}$ & $34.51_{1.53}$ & $\mathbf{39.57_{0.24}}$ & $38.10_{1.68}$\\
ANLI R3    & $41.26$ & $41.87$ & $41.32_{0.05}$ & $40.85_{0.36}$ & $\underline{42.20_{2.11}}$ & $35.98_{2.80}$ & $\mathbf{42.96_{0.42}}$ & $41.51_{2.33}$\\
CB         & $70.12$ & $70.12$ & $72.06_{0.14}$ & $71.31_{2.39}$ & $\underline{83.97_{5.40}}$ & $59.21_{5.65}$ & $76.55_{0.41}$ & $\mathbf{80.12_{1.61}}$\\
RTE        & $80.83$ & $80.97$ & $81.73_{0.49}$ & $\underline{82.47_{0.86}}$ & $79.51_{2.08}$ & $64.86_{10.29}$ & $81.71_{0.12}$ & $\mathbf{84.22_{1.02}}$\\
WSC        & $61.45$ & $61.06$ & $60.74_{0.31}$ & $63.30_{1.82}$ & $\underline{64.65_{1.79}}$ & $61.03_{3.86}$ & $61.47_{1.44}$ & $\mathbf{63.62_{1.68}}$\\
Winogrande & $59.94$ & $59.70$ & $59.46_{0.32}$ & $58.63_{0.70}$ & $\underline{59.76_{1.47}}$ & $53.22_{1.58}$ & $58.11_{0.26}$ & $\mathbf{59.59_{2.06}}$\\
COPA       & $90.02$ & $90.02$ & $90.51_{0.74}$ & $92.33_{0.39}$ & $\underline{92.54_{0.98}}$ & $82.82_{1.39}$ & $91.75_{0.42}$ & $\mathbf{93.50_{0.14}}$\\
HellaSwag  & $33.55$ & $33.52$ & $33.47_{0.20}$ & $37.28_{0.29}$ & $\underline{49.75_{4.98}}$ & $27.35_{2.28}$ & $33.09_{0.20}$ & $\mathbf{38.85_{5.54}}$\\
WiC        & $56.68$ & $56.13$ & $57.09_{0.40}$ & $58.86_{1.22}$ & $\underline{59.04_{1.09}}$ & $50.35_{0.89}$ & $57.03_{0.67}$ & $\mathbf{57.65_{1.18}}$\\
\midrule
Avg        & $57.60$ & $57.52$ & $57.82_{0.03}$ & $58.56_{0.20}$ & $\underline{61.73_{0.09}}$ & $51.28_{1.66}$ & $58.66_{0.35}$ & $\mathbf{60.12_{1.40}}$\\
\bottomrule
\end{tabular}
\caption{Main results (accuracy) on the test benchmark of different methods. Black-Box Tuning (BBT), Prompt Tuning (PT), Model Tuning (MT) tune the continuous prompt embedding or the full model for few-shot learning, while In-Context Learning (ICL), GRIPS~\cite{prasad2022grips} and our GPS do not update any parameter. We repeat the experiments of few-shot methods with 3 different data splits and report the average performance across all prompts. The number of subscript is the standard deviation across different data splits. T0, BBT, PT, MT, and ICL all use  the same manual prompt set, while GRIPS and GPS use their final searched prompt sets, where the number of prompts are the same in different prompts sets. \underline{Underlined} results are the best of few-shot parameter tuning methods and \textbf{bold} results are the best of few-shot parameter frozen methods. \dag: The original results from \citet{sanh2021multitask}. \ddag: The results we reproduced.}
\label{tab:main_results}
\end{table*}

\subsection{Implementation Details}
In practice, we assume there is only a few-shot validation set to conduct our experiments, which means, we will not tune any hyper-parameter in the method according to the performance on test set. Specifically, we set K as the number of initial prompts for each. It is a reasonable setup, because if the K is too low, the method may simply drop all prompts of low quality and keep the rest prompts as the final result. In the main experiment, we run the genetic prompt search for 6 steps. To generate diverse candidate prompts at each step, we perform top-p sampling, where the p is set as 0.9. Besides, we filter out all prompts that are the same as the existing prompts or do not have a valid input placeholder, such as "premise" in SuperGLUE CB.

\subsection{Main Results}

In this section, we compare \proposed with the aforementioned baselines on the 10 unseen tasks.


As shown in Table~\ref{tab:main_results}, \proposed outperforms other few-shot learning methods that do not require parameter updating. Compared to GRIPS, another discrete prompt search method, GPS wins 1.4 points on average, and the performance of GRIPS is close to that of PT while worse than MT. This reveals that the way to conduct prompt generation is important to obtaining the optimal prompts, and we also conduct further ablation studies on different prompt generation strategies in Sec~\ref{subsubsec:prompt_generation_strategies}.
Meanwhile, we find that the result of ICL with T0 is significantly worse than the zero-shot baseline, which differs from the results in previous works using GPT. We suppose this is because the prompts used in T0 at the multitask pretraining stage are quite different from the ICL demonstrations with labeled examples.
Even compared with parameter-efficient tuning methods like black-box tuning and prompt tuning, \proposed also achieves considerably better performance.

\begin{table}[t]
\small
\centering
\begin{tabular}
{cm{0.25in}<{\centering}m{0.25in}<{\centering}m{0.25in}<{\centering}m{0.3in}<{\centering}m{0.3in}<{\centering}}
\toprule
Dataset & T0\dag & BT & Cloze & SC (GPT2) & SC (T5LM) \\
\midrule
ANLI R1    &43.16 & 44.95 & 42.10 & 44.64 & \textbf{46.47}\\
ANLI R2    &38.68 & \textbf{40.13} & 38.95 & 39.14 & 39.91\\
ANLI R3    &41.87 & 42.83 & 42.00 & 42.75 & \textbf{43.01}\\
CB         &70.12 & 79.38 & 73.21 & 79.71 & \textbf{80.00}\\
RTE        &80.97 & 82.60 & 82.64 & 82.53 & \textbf{83.86}\\
WSC        &61.06 & 65.38 & 64.38 & 63.65 & \textbf{65.48}\\
Winogrande &59.70 & 61.09 & 53.50 & 59.76 & \textbf{61.96}\\
COPA       &90.02 & 93.12 & 89.77 & 93.31 & \textbf{93.43}\\
HellaSwag  &33.52 & 36.34 & 33.83 & 36.72 & \textbf{44.29}\\
WiC        &56.13 & \textbf{60.72} & 56.15 & 57.91 & 58.82\\
\midrule
Avg        &57.52 & 60.65 & 57.65 & 60.01 & \textbf{61.72}\\
\bottomrule
\end{tabular}
\caption{Ablation results with different prompt generation strategies including Cloze, Back Translation (BT) and Sentence Continuation (SC). For SC, we include two different pretrained models, GPT2 and T5LM. \dag: The results we reproduced.}
\label{tab:ablation_prompt_generation_strategies}
\end{table}


\subsection{Ablation Study}
In this section, we conduct several ablation experiments on various hyper-parameters. To control experimental variables, we explore the effect of each hyper-parameter while keeping the other hyper-parameters fixed as the default value. 

\subsubsection{Prompt Generation Strategies}
\label{subsubsec:prompt_generation_strategies}

In Table~\ref{tab:ablation_prompt_generation_strategies}, we compare different automatic prompt generation strategies, including Cloze, Back Translation (BT), and Sentence Continuation (SC) as described in section~\ref{subsection:prompt_generation}. Among all the prompt generation strategies, SC with T5LM obtains the best results. And SC with T5LM shows that it outperforms SC with GPT2 by a significant margin. This reveals that the pretrained model for SC is vital to obtaining the optimal prompts, and larger language models usually generate better prompts. 

BT achieves good results and has the best score on ANLI R2 and WiC. Cloze does not work well and its overall performance is no better than the zero-shot baseline. The results suggest that these simple strategies cannot provide enough variance of prompts for search, and we will discuss this with a few examples in section~\ref{sub:case_study}.

\begin{table*}[t]
\small
\centering
\begin{tabular}{m{1.4cm}<{\centering}p{12cm}m{1.0cm}<{\centering}}
\toprule
\textbf{Task} & \textbf{Generated Prompts} & \textbf{Metric} \\
\midrule
\multirow{4}{*}{\makecell[c]{\\ Hellaswag}} & \textbf{Origin}: If a description of a situation begins like this: \{\{ ctx \}\}... Then how does it continue? & 34.00 \\
~ & \textbf{BT}: If the description of a situation begins like this: \{\{ ctx \}\}... Then how will it continue? & 34.74 \\
~ & \textbf{SC(GPT2)}: - & 34.00 \\
~ & \textbf{SC(T5LM)}:  If a description of a situation begins like this: \{\{ ctx \}\}... then what is the most likely thing to happen next? & \textbf{47.63} \\
\midrule
\multirow{4}{*}{\makecell[c]{\\SuperGLUE \\ COPA}} & \textbf{Origin}: Select the most plausible \{\% if question == "cause" \%\} cause: \{\% else \%\} effect: \{\% endif \%\} & 93.00\\
~ & \textbf{BT}: Select the most believable \{\% if question == "cause" \%\} cause: \{\% else \%\} effect: \{\% endif \%\} & 95.00 \\
~ & \textbf{SC(GPT2)}: What is the most plausible \{\% if question == "cause" \%\} cause: \{\% else \%\} effect:\{\% endif \%\} & 95.00 \\
~ & \textbf{SC(T5LM)}: Select the most agreeable \{\% if question == "cause" \%\} cause: \{\% else \%\} effect:\{\% endif \%\} & \textbf{96.00} \\
\midrule
\multirow{4}{*}{\makecell[c]{\\SuperGLUE \\ CB}} & \textbf{Origin}: \{\{premise\}\} Are we justified in saying that "\{\{hypothesis\}\}"? Yes, no, or maybe? & 78.57 \\
~ & \textbf{BT}: \{\{premise\}\} Do we have reason to say this "\{\{hypothesis\}\}"? Yes, no, or maybe? & 83.93 \\
~ & \textbf{SC(GPT2)}: \{\{premise\}\} Are we justified in believing that "\{\{hypothesis\}\}"? Yes, no, or maybe? & 80.36 \\
~ & \textbf{SC(T5LM)}: \{\{premise\}\} If we were justified, would we think that it is the case that we are justified in saying that "\{\{hypothesis\}\}"? Yes, no, or maybe? & \textbf{83.93} \\

\midrule
\multirow{4}{*}{\makecell[c]{\\SuperGLUE \\ WSC}} & \textbf{Origin}: Passage: \{\{text\}\} Question: In the passage above, does the pronoun "\{\{span2\_text\}\}" refer to \{\{span1\_text\}\}? Answer: & 60.58 \\
~ & \textbf{BT}: Passage: \{\{ text \}\} Question: in the paragraph above, does the pronoun "\{\{ span2\_text \}\}" refer to \{\{ span1\_text \}\}? Answer: & 67.31 \\
~ & \textbf{SC(GPT2)}: - & 60.58 \\
~ & \textbf{SC(T5LM)}: Passage: \{\{ text \}\} Question: does the pronoun "\{\{ span2\_text \}\}" refer to the person of \{\{ span1\_text \}\}? Answer: & \textbf{70.19} \\

\bottomrule
\end{tabular}
\caption{Illustration of prompts generated by GPS. “-” indicates when the score of the original prompt is better than the generated prompts, and GPS will keep the original prompt as the final result.}
\label{tab:en-gps}
\end{table*}

\subsubsection{The Size of Validation Set}


\begin{figure}[t]
    \centering
    \includegraphics[width=0.45\textwidth]{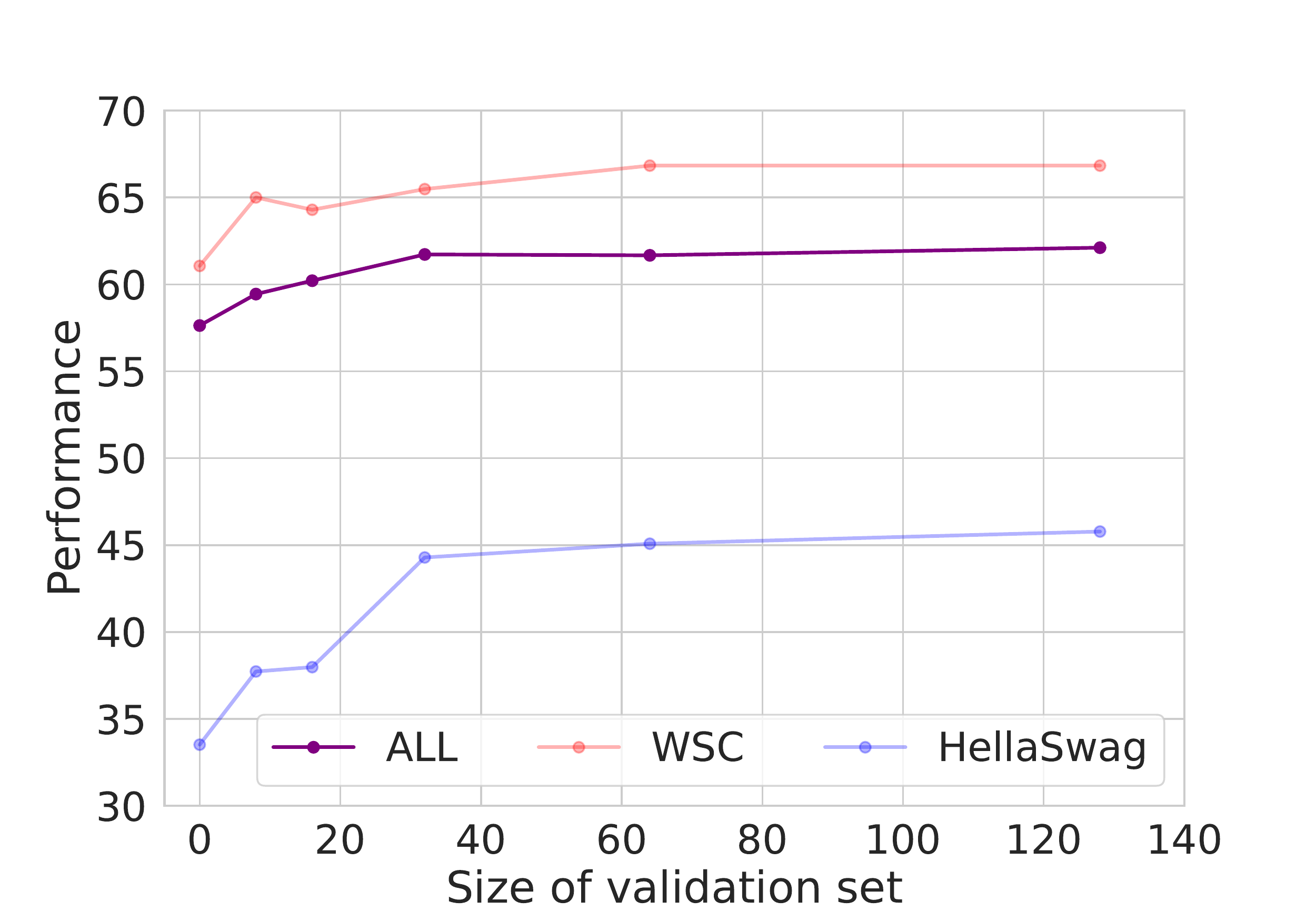}
    \caption{Ablation results on the size of the validation set. The performance with an empty validation set is zero-shot T0.
}
    \label{fig:ablation_val_size}
\end{figure}

The validation set plays an important role in GPS for scoring each prompt, and its size matters to give credible feedback for the prompt selection. 
The results of GPS on validation sets of different sizes from 8 to 128 are presented in Fig.~\ref{fig:ablation_val_size}. Generally, the gain of prompt search rises with more validation samples. Most datasets follow this trend, such as WSC and HellaSwag. Although more examples lead to further improvement, we set the default size of validation set to 32 in our experiments because we focus on the few-shot scenario with limited labeled points. The size of the prompt pool might be important as well for continuous improvement, especially with a large validation set. Here we set the prompt pool size to be 30 in consideration of the computational costs. On the other hand, GPS is still much better than manual prompts when there is only 8 examples for validation.

\subsubsection{The Number of Prompt Search Iterations}
\begin{figure}[t]
    \centering
    \includegraphics[width=0.45\textwidth]{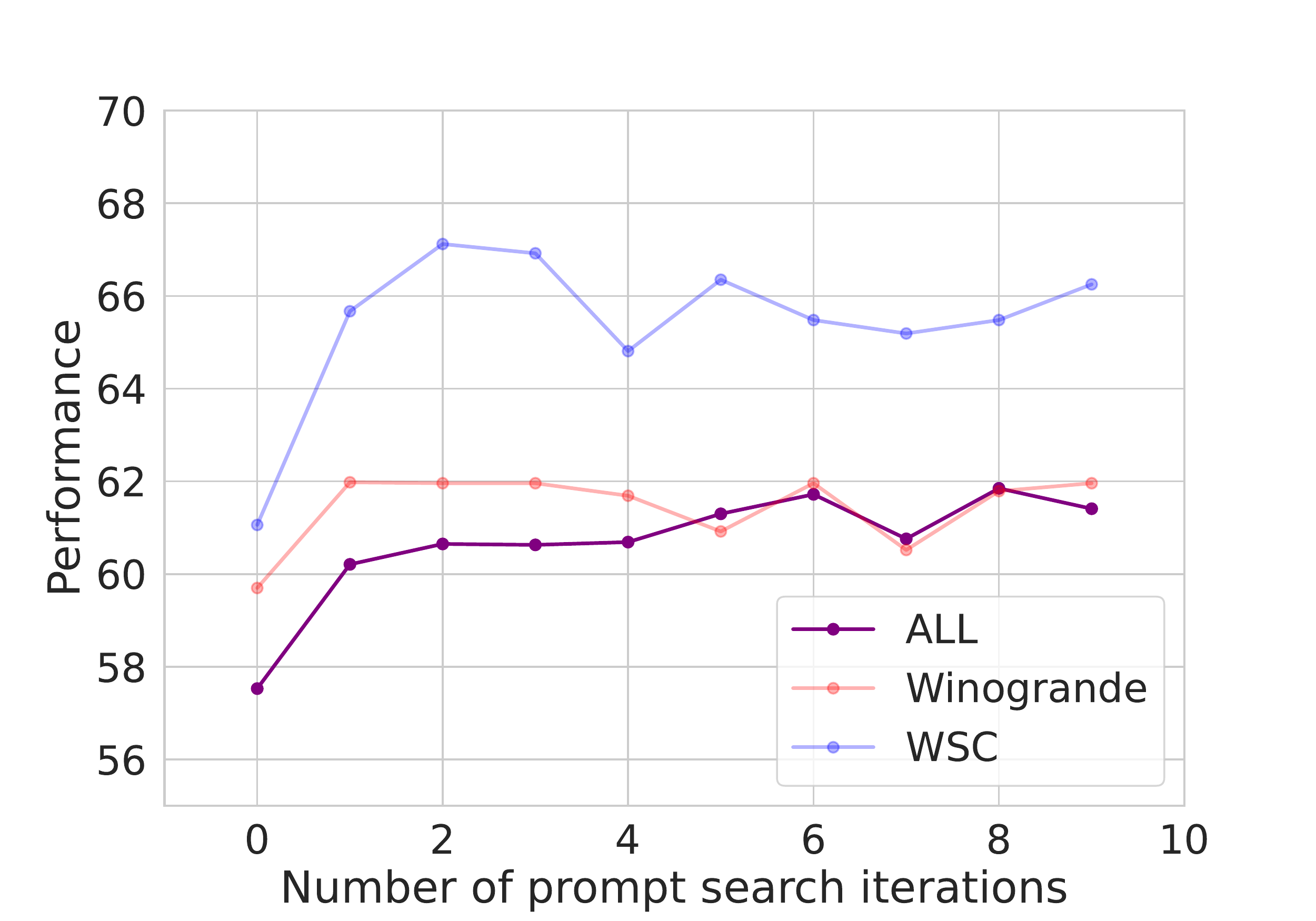}
    \caption{Ablation results on the number of prompt search iterations. The performance with 0 iterations is zero-shot T0.
}
    \label{fig:ablation_num_iter}
\end{figure}

Another critical hyperparameter is the number of iterations for genetic prompt search. We experimented with up to 9 prompt sesarch iterations and the results are given in Fig.~\ref{fig:ablation_num_iter}. It can be seen that the performances on some datasets such as WSC and Winogrande achieve the best at an early iteration. However, the overall performance on all the datasets improves on more search iterations. The default iteration number is set to 6 in the paper for the trade-off between the performance and costs. 

\begin{table*}[t]
\small
\centering
\begin{tabular}{cm{0.8in}<{\centering}m{0.8in}<{\centering}m{0.8in}<{\centering}m{0.8in}<{\centering}}
\toprule
Methods & Serving Efficiency & Tunable Parameters & Performance & Computation Cost$^\dag$\\
\midrule
Model Tuning         & \xmark & 100\%         & 61.73 & 11.1x \\
Prompt Tuning        & \cmark & $\sim$ 0.01\% & 58.56 & 11.1x \\
Black-Box Tuning     & \cmark & $\sim$ 0.001\%  & 57.82 & 9.3x \\
In-Context Learning  & \xmark$^\ddag$ & 0\%           & 51.28 & 0x \\
our GPS              & \cmark & 0\%           & 60.12 & 1.0x \\
\bottomrule
\end{tabular}
\caption{Overall comparison of different few-shot learning methods on serving efficiency, tunable parameters, performance and computation cost. $\dag$: Computation cost here refers to the combined cost of training and prompt search. $\ddag$: In-context learning uses a long sequence length to concatenate examples, which is expensive for inference.}
\label{tab:overall_comparison}
\end{table*}


\subsection{Case Study} 
\label{sub:case_study}
In Table \ref{tab:en-gps}, we present cases of prompts selected by GPS using different strategies. As we can see, GPS modifies the original prompts to optimize performance for unseen tasks while not changing the major meanings.
However, using simple strategies like back translation can only provide minor prompt modifications, while SC with T5LM shows larger prompt modifications and dramatically better performance. 
For example, in WSC, GPS firstly removes the less informative adverb ``In the passage above'', and then adds a hint ``the person of'' to help the model navigate the answer. This modification obtains a significant 10 points gain compared to the original prompt. 
Overall, Table~\ref{tab:en-gps} illustrates the necessity and effectiveness of \proposed, especially when SC is used on a large language model.

\subsection{Overall Comparison of Different Few-Shot Methods}

Table~\ref{tab:overall_comparison} compares different methods on serving efficiency, tunable parameters, performance, and computational cost.

\textbf{Serving Efficiency.} MT lacks serving efficiency due to the huge storage cost to store the full model for each new task. Although ICL does not have any tunable parameter, the long sequence length makes it expensive for inference, especially when the number of demonstrations is large. PT, BBT and GPS have few or zero tunable parameters, and thus they are cheaper for deployment.

\textbf{Tunable Parameters.} MT needs to tune the full model for each task, and it requires large resources because the commonly-used optimizers such as Adam require to store extra momentum and variance terms. PT and BBT only tune the prompt embeddings and are more efficient.
ICL and GPS are the most parameter-efficient as they require no parameter updating.

\textbf{Performance.} GPS has the second best performance even though all the model parameters are frozen. MT is better than PT, and PT is better than BBT, which is different from the results given in~\citet{Sun2021blackbox}. We suppose the reason is that we use different pretrained models and test datasets, and a more strict setting where only 16 examples are used for the train and dev set, respectively.

\textbf{Computational Cost.} We consider the computational cost as the number of equivalent forward passes during the training or the prompt search stage. 
The training batch size for MT and PT is 4 and the total training step is 4000. 
We estimate the computational cost for each backward pass as two forward passes. 
The number of the manual prompts and the topK for prompt selection are both 5.
For BBT, the training iteration is 500 and the prompt number is 5, and we omit the cost of CMA-ES. For GPS, we consider 6 search iterations and a prompt pool size of 30, the cost for generating one prompt is estimated as two forward passes.
In total, the equivalent computation costs for MT and PT are 48000 forward passes, and they are 40000 and 4320 for BBT and GPS, respectively.

Overall, our GPS has good serving-efficiency, low computation cost, does not have any tunable parameter, and still achieves the best performance. It can be a promising option for a large-scale NLP production system to improve the performance of pretrained language models with only limited labeled examples.

\section{Conclusions}
In this paper, we propose \proposed, an automatic prompt search method based on genetic algorithm for better few-shot learning. We compare different approaches on 10 datasets with only 32 labeled examples available. \proposed outperforms not only the manual prompt baseline, but also other parameter-efficient few-shot learning methods. Extensive experiments verified the effectiveness of the proposed \proposed.

\section{Limitations}

We show that Genetic Prompt Search is an efficient few-shot learning approach with competitive performance as well as low cost.
Our results have a few limitations, however, and it is possible that few-shot performance could be further improved by studying those problems in the future.
Specifically, 
1) We conduct experiments on the T0 benchmark with 10 test datasets. It is not clear how our method performs on other datasets.
2) We only compare different methods under the few-shot setting with 32 examples in total. Conclusions regarding the performance of different methods might be different with more labeled examples. For example, if it is possible to get hundreds of or even thousands of training examples, tuning-based methods might achieve much better and more stable performance.
3) Although GPS is able to find better prompts automatically, it is still not clear why these prompts work better. Further research on the mechanism of prompting on large-scale language models can help us understand what kind of prompt works and how to design optimal prompts.
We hope our results could encourage future work on addressing these limitations to further explore the potential of few-shot learning.

\bibliography{custom}

\begin{thebibliography}{29}
\expandafter\ifx\csname natexlab\endcsname\relax\def\natexlab#1{#1}\fi

\bibitem[{Ben~Zaken et~al.(2022)Ben~Zaken, Goldberg, and
  Ravfogel}]{ben-zaken-etal-2022-bitfit}
Elad Ben~Zaken, Yoav Goldberg, and Shauli Ravfogel. 2022.
\newblock \href {https://doi.org/10.18653/v1/2022.acl-short.1} {{B}it{F}it:
  Simple parameter-efficient fine-tuning for transformer-based masked
  language-models}.
\newblock In \emph{Proceedings of the 60th Annual Meeting of the Association
  for Computational Linguistics (Volume 2: Short Papers)}, pages 1--9, Dublin,
  Ireland. Association for Computational Linguistics.

\bibitem[{Brown et~al.(2020)Brown, Mann, Ryder, Subbiah, Kaplan, Dhariwal,
  Neelakantan, Shyam, Sastry, Askell, Agarwal, Herbert-Voss, Krueger, Henighan,
  Child, Ramesh, Ziegler, Wu, Winter, Hesse, Chen, Sigler, Litwin, Gray, Chess,
  Clark, Berner, McCandlish, Radford, Sutskever, and Amodei}]{NEURIPS2020_gpt3}
Tom Brown, Benjamin Mann, Nick Ryder, Melanie Subbiah, Jared~D Kaplan, Prafulla
  Dhariwal, Arvind Neelakantan, Pranav Shyam, Girish Sastry, Amanda Askell,
  Sandhini Agarwal, Ariel Herbert-Voss, Gretchen Krueger, Tom Henighan, Rewon
  Child, Aditya Ramesh, Daniel Ziegler, Jeffrey Wu, Clemens Winter, Chris
  Hesse, Mark Chen, Eric Sigler, Mateusz Litwin, Scott Gray, Benjamin Chess,
  Jack Clark, Christopher Berner, Sam McCandlish, Alec Radford, Ilya Sutskever,
  and Dario Amodei. 2020.
\newblock \href
  {https://proceedings.neurips.cc/paper/2020/file/1457c0d6bfcb4967418bfb8ac142f64a-Paper.pdf}
  {Language models are few-shot learners}.
\newblock In \emph{Advances in Neural Information Processing Systems},
  volume~33, pages 1877--1901. Curran Associates, Inc.

\bibitem[{Devlin et~al.(2019)Devlin, Chang, Lee, and
  Toutanova}]{devlin-etal-2019-bert}
Jacob Devlin, Ming-Wei Chang, Kenton Lee, and Kristina Toutanova. 2019.
\newblock \href {https://doi.org/10.18653/v1/N19-1423} {{BERT}: Pre-training of
  deep bidirectional transformers for language understanding}.
\newblock In \emph{Proceedings of the 2019 Conference of the North {A}merican
  Chapter of the Association for Computational Linguistics: Human Language
  Technologies, Volume 1 (Long and Short Papers)}, pages 4171--4186,
  Minneapolis, Minnesota. Association for Computational Linguistics.

\bibitem[{Gao et~al.(2021{\natexlab{a}})Gao, Tow, Biderman, Black, DiPofi,
  Foster, Golding, Hsu, McDonell, Muennighoff, Phang, Reynolds, Tang, Thite,
  Wang, Wang, and Zou}]{eval-harness}
Leo Gao, Jonathan Tow, Stella Biderman, Sid Black, Anthony DiPofi, Charles
  Foster, Laurence Golding, Jeffrey Hsu, Kyle McDonell, Niklas Muennighoff,
  Jason Phang, Laria Reynolds, Eric Tang, Anish Thite, Ben Wang, Kevin Wang,
  and Andy Zou. 2021{\natexlab{a}}.
\newblock \href {https://doi.org/10.5281/zenodo.5371628} {A framework for
  few-shot language model evaluation}.

\bibitem[{Gao et~al.(2021{\natexlab{b}})Gao, Fisch, and
  Chen}]{gao-etal-2021-making}
Tianyu Gao, Adam Fisch, and Danqi Chen. 2021{\natexlab{b}}.
\newblock \href {https://doi.org/10.18653/v1/2021.acl-long.295} {Making
  pre-trained language models better few-shot learners}.
\newblock In \emph{Proceedings of the 59th Annual Meeting of the Association
  for Computational Linguistics and the 11th International Joint Conference on
  Natural Language Processing (Volume 1: Long Papers)}, pages 3816--3830,
  Online. Association for Computational Linguistics.

\bibitem[{Han et~al.(2021)Han, Zhao, Ding, Liu, and Sun}]{han2021ptr}
Xu~Han, Weilin Zhao, Ning Ding, Zhiyuan Liu, and Maosong Sun. 2021.
\newblock \href {http://arxiv.org/abs/2105.11259} {Ptr: Prompt tuning with
  rules for text classification}.

\bibitem[{Houlsby et~al.(2019)Houlsby, Giurgiu, Jastrzebski, Morrone,
  De~Laroussilhe, Gesmundo, Attariyan, and Gelly}]{pmlr-v97-houlsby19a}
Neil Houlsby, Andrei Giurgiu, Stanislaw Jastrzebski, Bruna Morrone, Quentin
  De~Laroussilhe, Andrea Gesmundo, Mona Attariyan, and Sylvain Gelly. 2019.
\newblock \href {https://proceedings.mlr.press/v97/houlsby19a.html}
  {Parameter-efficient transfer learning for {NLP}}.
\newblock In \emph{Proceedings of the 36th International Conference on Machine
  Learning}, volume~97 of \emph{Proceedings of Machine Learning Research},
  pages 2790--2799. PMLR.

\bibitem[{Hu et~al.(2021)Hu, Shen, Wallis, Allen-Zhu, Li, Wang, Wang, and
  Chen}]{Hu2021LoRA}
Edward~J. Hu, Yelong Shen, Phillip Wallis, Zeyuan Allen-Zhu, Yuanzhi Li, Shean
  Wang, Lu~Wang, and Weizhu Chen. 2021.
\newblock \href {https://doi.org/10.48550/ARXIV.2106.09685} {Lora: Low-rank
  adaptation of large language models}.

\bibitem[{Le~Scao and Rush(2021)}]{le-scao-rush-2021-many}
Teven Le~Scao and Alexander Rush. 2021.
\newblock \href {https://doi.org/10.18653/v1/2021.naacl-main.208} {How many
  data points is a prompt worth?}
\newblock In \emph{Proceedings of the 2021 Conference of the North American
  Chapter of the Association for Computational Linguistics: Human Language
  Technologies}, pages 2627--2636, Online. Association for Computational
  Linguistics.

\bibitem[{Lester et~al.(2021)Lester, Al-Rfou, and Constant}]{lester2021power}
Brian Lester, Rami Al-Rfou, and Noah Constant. 2021.
\newblock \href {http://arxiv.org/abs/2104.08691} {The power of scale for
  parameter-efficient prompt tuning}.

\bibitem[{Liu et~al.(2021{\natexlab{a}})Liu, Yuan, Fu, Jiang, Hayashi, and
  Neubig}]{liu2021pre}
Pengfei Liu, Weizhe Yuan, Jinlan Fu, Zhengbao Jiang, Hiroaki Hayashi, and
  Graham Neubig. 2021{\natexlab{a}}.
\newblock Pre-train, prompt, and predict: A systematic survey of prompting
  methods in natural language processing.
\newblock \emph{arXiv preprint arXiv:2107.13586}.

\bibitem[{Liu et~al.(2021{\natexlab{b}})Liu, Zheng, Du, Ding, Qian, Yang, and
  Tang}]{liu2021gpt}
Xiao Liu, Yanan Zheng, Zhengxiao Du, Ming Ding, Yujie Qian, Zhilin Yang, and
  Jie Tang. 2021{\natexlab{b}}.
\newblock \href {http://arxiv.org/abs/2103.10385} {Gpt understands, too}.

\bibitem[{Mishra et~al.(2021)Mishra, Khashabi, Baral, Choi, and
  Hajishirzi}]{mishra2021reframing}
Swaroop Mishra, Daniel Khashabi, Chitta Baral, Yejin Choi, and Hannaneh
  Hajishirzi. 2021.
\newblock \href {http://arxiv.org/abs/2109.07830} {Reframing instructional
  prompts to gptk's language}.

\bibitem[{Mitchell(1980)}]{mitchell80}
T.~M. Mitchell. 1980.
\newblock The need for biases in learning generalizations.
\newblock Technical report, Computer Science Department, Rutgers University,
  New Brunswick, MA.

\bibitem[{Ouyang et~al.(2022)Ouyang, Wu, Jiang, Almeida, Wainwright, Mishkin,
  Zhang, Agarwal, Slama, Ray, Schulman, Hilton, Kelton, Miller, Simens, Askell,
  Welinder, Christiano, Leike, and Lowe}]{Ouyang2022InstructGPT}
Long Ouyang, Jeff Wu, Xu~Jiang, Diogo Almeida, Carroll~L. Wainwright, Pamela
  Mishkin, Chong Zhang, Sandhini Agarwal, Katarina Slama, Alex Ray, John
  Schulman, Jacob Hilton, Fraser Kelton, Luke Miller, Maddie Simens, Amanda
  Askell, Peter Welinder, Paul Christiano, Jan Leike, and Ryan Lowe. 2022.
\newblock \href {https://doi.org/10.48550/ARXIV.2203.02155} {Training language
  models to follow instructions with human feedback}.

\bibitem[{Perez et~al.(2021)Perez, Kiela, and Cho}]{perez2021true}
Ethan Perez, Douwe Kiela, and Kyunghyun Cho. 2021.
\newblock \href
  {https://proceedings.neurips.cc/paper/2021/hash/5c04925674920eb58467fb52ce4ef728-Abstract.html}
  {True few-shot learning with language models}.
\newblock In \emph{Advances in Neural Information Processing Systems 34: Annual
  Conference on Neural Information Processing Systems 2021, NeurIPS 2021,
  December 6-14, 2021, virtual}, pages 11054--11070.

\bibitem[{Prasad et~al.(2022)Prasad, Hase, Zhou, and Bansal}]{prasad2022grips}
Archiki Prasad, Peter Hase, Xiang Zhou, and Mohit Bansal. 2022.
\newblock Grips: Gradient-free, edit-based instruction search for prompting
  large language models.
\newblock \emph{arXiv preprint arXiv:2203.07281}.

\bibitem[{Radford et~al.(2018)Radford, Wu, Child, Luan, Amodei, and
  Sutskever}]{radf2019lmmultitask}
Alec Radford, Jeffrey Wu, Rewon Child, David Luan, Dario Amodei, and Ilya
  Sutskever. 2018.
\newblock \href
  {https://d4mucfpksywv.cloudfront.net/better-language-models/language-models.pdf}
  {Language models are unsupervised multitask learners}.

\bibitem[{Raffel et~al.(2020)Raffel, Shazeer, Roberts, Lee, Narang, Matena,
  Zhou, Li, and Liu}]{JMLR:v21:20-074}
Colin Raffel, Noam Shazeer, Adam Roberts, Katherine Lee, Sharan Narang, Michael
  Matena, Yanqi Zhou, Wei Li, and Peter~J. Liu. 2020.
\newblock \href {http://jmlr.org/papers/v21/20-074.html} {Exploring the limits
  of transfer learning with a unified text-to-text transformer}.
\newblock \emph{Journal of Machine Learning Research}, 21(140):1--67.

\bibitem[{Sanh et~al.(2021)Sanh, Webson, Raffel, Bach, Sutawika, Alyafeai,
  Chaffin, Stiegler, Scao, Raja, Dey, Bari, Xu, Thakker, Sharma, Szczechla,
  Kim, Chhablani, Nayak, Datta, Chang, Jiang, Wang, Manica, Shen, Yong, Pandey,
  Bawden, Wang, Neeraj, Rozen, Sharma, Santilli, Fevry, Fries, Teehan,
  Biderman, Gao, Bers, Wolf, and Rush}]{sanh2021multitask}
Victor Sanh, Albert Webson, Colin Raffel, Stephen~H. Bach, Lintang Sutawika,
  Zaid Alyafeai, Antoine Chaffin, Arnaud Stiegler, Teven~Le Scao, Arun Raja,
  Manan Dey, M~Saiful Bari, Canwen Xu, Urmish Thakker, Shanya~Sharma Sharma,
  Eliza Szczechla, Taewoon Kim, Gunjan Chhablani, Nihal Nayak, Debajyoti Datta,
  Jonathan Chang, Mike Tian-Jian Jiang, Han Wang, Matteo Manica, Sheng Shen,
  Zheng~Xin Yong, Harshit Pandey, Rachel Bawden, Thomas Wang, Trishala Neeraj,
  Jos Rozen, Abheesht Sharma, Andrea Santilli, Thibault Fevry, Jason~Alan
  Fries, Ryan Teehan, Stella Biderman, Leo Gao, Tali Bers, Thomas Wolf, and
  Alexander~M. Rush. 2021.
\newblock \href {http://arxiv.org/abs/2110.08207} {Multitask prompted training
  enables zero-shot task generalization}.

\bibitem[{Schick and Sch{\"u}tze(2021)}]{schick-schutze-2021-exploiting}
Timo Schick and Hinrich Sch{\"u}tze. 2021.
\newblock \href {https://aclanthology.org/2021.eacl-main.20} {Exploiting
  cloze-questions for few-shot text classification and natural language
  inference}.
\newblock In \emph{Proceedings of the 16th Conference of the European Chapter
  of the Association for Computational Linguistics: Main Volume}, pages
  255--269, Online. Association for Computational Linguistics.

\bibitem[{Schick and Schütze(2021)}]{schick2020generating}
Timo Schick and Hinrich Schütze. 2021.
\newblock \href {https://arxiv.org/abs/2104.07540} {Generating datasets with
  pretrained language models}.
\newblock \emph{Computing Research Repository}, arXiv:2104.07540.

\bibitem[{Shin et~al.(2020)Shin, Razeghi, Logan~IV, Wallace, and
  Singh}]{shin-etal-2020-autoprompt}
Taylor Shin, Yasaman Razeghi, Robert~L. Logan~IV, Eric Wallace, and Sameer
  Singh. 2020.
\newblock \href {https://doi.org/10.18653/v1/2020.emnlp-main.346}
  {{A}uto{P}rompt: {E}liciting {K}nowledge from {L}anguage {M}odels with
  {A}utomatically {G}enerated {P}rompts}.
\newblock In \emph{Proceedings of the 2020 Conference on Empirical Methods in
  Natural Language Processing (EMNLP)}, pages 4222--4235, Online. Association
  for Computational Linguistics.

\bibitem[{Sun et~al.(2022{\natexlab{a}})Sun, Shao, Qian, Huang, and
  Qiu}]{Sun2021blackbox}
Tianxiang Sun, Yunfan Shao, Hong Qian, Xuanjing Huang, and Xipeng Qiu.
  2022{\natexlab{a}}.
\newblock \href {https://doi.org/10.48550/ARXIV.2201.03514} {Black-box tuning
  for language-model-as-a-service}.

\bibitem[{Sun et~al.(2022{\natexlab{b}})Sun, Shao, Qian, Huang, and
  Qiu}]{sun2022black}
Tianxiang Sun, Yunfan Shao, Hong Qian, Xuanjing Huang, and Xipeng Qiu.
  2022{\natexlab{b}}.
\newblock Black-box tuning for language-model-as-a-service.
\newblock \emph{arXiv preprint arXiv:2201.03514}.

\bibitem[{Wang et~al.(2022)Wang, Roberts, Hesslow, Scao, Chung, Beltagy,
  Launay, and Raffel}]{wang2022language}
Thomas Wang, Adam Roberts, Daniel Hesslow, Teven~Le Scao, Hyung~Won Chung,
  Iz~Beltagy, Julien Launay, and Colin Raffel. 2022.
\newblock What language model architecture and pretraining objective work best
  for zero-shot generalization?
\newblock \emph{arXiv preprint arXiv:2204.05832}.

\bibitem[{Webson and Pavlick(2021)}]{webson2021prompt}
Albert Webson and Ellie Pavlick. 2021.
\newblock Do prompt-based models really understand the meaning of their
  prompts?
\newblock \emph{arXiv preprint arXiv:2109.01247}.

\bibitem[{Wei et~al.(2021)Wei, Bosma, Zhao, Guu, Yu, Lester, Du, Dai, and
  Le}]{wei2021finetuned}
Jason Wei, Maarten Bosma, Vincent~Y. Zhao, Kelvin Guu, Adams~Wei Yu, Brian
  Lester, Nan Du, Andrew~M. Dai, and Quoc~V. Le. 2021.
\newblock \href {http://arxiv.org/abs/2109.01652} {Finetuned language models
  are zero-shot learners}.

\bibitem[{Yang et~al.(2019)Yang, Dai, Yang, Carbonell, Salakhutdinov, and
  Le}]{yang2019xlnet}
Zhilin Yang, Zihang Dai, Yiming Yang, Jaime Carbonell, Russ~R Salakhutdinov,
  and Quoc~V Le. 2019.
\newblock Xlnet: Generalized autoregressive pretraining for language
  understanding.
\newblock \emph{Advances in neural information processing systems}, 32.

\end{thebibliography}
\bibliographystyle{acl_natbib}


\end{document}